\documentclass[times, review, 10pt]{elsarticle}
\makeatletter
\def\ps@pprintTitle{%
  \let\@oddhead\@empty
  \let\@evenhead\@empty
  \let\@oddfoot\@empty
  \let\@evenfoot\@empty
}
\makeatother
\usepackage{amssymb}

\usepackage{amsmath}
\usepackage[symbol]{footmisc}
\usepackage{array}
\usepackage{booktabs}
\usepackage{threeparttable}
\usepackage{multirow}
\usepackage{booktabs}
\usepackage[table]{xcolor}
\usepackage{algorithm}
\usepackage{algorithmic}
\usepackage{float}
\usepackage{booktabs}
\usepackage[table]{xcolor}
\usepackage{multirow}
\usepackage{array}

\begin{document}

\begin{frontmatter}



\title{CoRe-ECG: Advancing Self-Supervised Representation Learning for 12-Lead ECG via Contrastive and Reconstructive Synergy}


\author{
Zehao Qin$^{a,}$\footnote[2]{\label{note1}Equal Contribution.},
Xiaojian Lin$^{d,}$\footref{note1},
Ping Zhang$^{d,}$,
Hongliang Wu$^{e,}$,
Xinkang Wang$^{f,}$,
Guangling Liu$^{a,}$,
Bo Chen$^{d,}$\footnote[1]{\label{note2}Correspondence: Bo Chen (cba04570@btch.edu.cn); Wenming Yang (yang.wenming@sz.tsinghua.edu.cn); Guijin Wang (wangguijin@tsinghua.edu.cn)
},
Wenming Yang$^{a,}$\footref{note2},
Guijin Wang$^{b,c,}$\footref{note2}
} %% Author name

% \affiliation{organization={},%Department and Organization

\affiliation[1]{organization={Department of Electronic Engineering, Shenzhen International Graduate School, Tsinghua University},
                city={Shenzhen},
                citysep={}, 
                postcode={518071}, 
                country={PR China}}

\affiliation[2]{organization={Department of Electronic Engineering, Tsinghua University},
                city={Beijing},
                citysep={}, % Uncomment if no comma needed between city and postcode
                postcode={100084}, 
                country={PR China}}

\affiliation[3]{organization={Shanghai AI Laboratory},
                city={Shanghai},
                postcode={200232},
                country={PR China}}

\affiliation[4]{organization={Department of Cardiology, Beijing Tsinghua Changgung Hospital, School of Clinical Medicine, Tsinghua Medicine, Tsinghua University},
                city={Beijing},
                postcode={102218},
                country={PR China}}

\affiliation[5]{organization={Department of Anesthesiology, National Cancer Center/National Clinical Research Center for Cancer/Cancer Hospital, Chinese Academy of Medical Sciences and Peking Union Medical College},
                city={Beijing},
                postcode={100021},
                country={PR China}}

\affiliation[6]{organization={Department of Electrocardiographic Diagnosis, Fuzhou University Affiliated Provincial Hospital},
                city={Fuzhou},
                postcode={350001},
                country={PR China}}

% \cortext[cor1]{Equal Contribution.}
% \cortext[cor2]{Corresponding author.
% \textit{E-mail address}: wangguijin@tsinghua.edu.cn (G. Wang).}

%% Abstract
\begin{abstract}
% Robotic grasping is a primitive skill for complex tasks and is fundamental to intelligence. For general 6-DoF grasping, most previous methods directly extract scene-level semantic or geometric information, while few of them consider the suitability for various downstream applications, such as target-oriented grasping. Addressing this issue, we rethink 6-DoF grasp detection from a grasp-centric view and propose a versatile grasp framework capable of handling both scene-level and target-oriented grasping. Our framework, \textit{FlexLoG}, is composed of a \textit{Flex}ible Guidance Module and a \textit{Lo}cal \textit{G}rasp model. Specifically, the Flexible Guidance Module is compatible with both global (e.g., grasp heatmap) and local (e.g., visual language grounding) guidance, enabling the generation of high-quality grasps across various tasks. The Local Grasp model focuses on object-agnostic regional points and predicts grasps locally and intently. Experiment results reveal that our framework achieves over 18\% and 23\% improvement on unseen splits of the GraspNet-1Billion Dataset. Constructed on a digital twin of the dataset in simulation, we evaluated all methods in the simulation benchmark to further validate the performance and align it with the dataset results. Furthermore, real-world robotic tests in distinct settings yield a over 90\% success rate.
Accurate interpretation of electrocardiogram (ECG) remains challenging due to the scarcity of labeled data and the high cost of expert annotation. Self-supervised learning (SSL) offers a promising solution by enabling models to learn expressive representations from unlabeled signals. Existing ECG SSL methods typically rely on either contrastive learning or reconstructive learning. However, each approach in isolation provides limited supervisory signals and suffers from additional limitations, including non-physiological distortions introduced by naive augmentations and trivial correlations across multiple leads that models may exploit as shortcuts. In this work, we propose CoRe-ECG, a unified contrastive and reconstructive pretraining paradigm that establishes a synergistic interaction between global semantic modeling and local structural learning. CoRe-ECG aligns global representations during reconstruction, enabling instance-level discriminative signals to guide local waveform recovery. To further enhance pretraining, we introduce Frequency Dynamic Augmentation (FDA) to adaptively perturb ECG signals based on their frequency-domain importance, and Spatio-Temporal Dual Masking (STDM) to break linear dependencies across leads, increasing the difficulty of reconstructive tasks. Our method achieves state-of-the-art performance across multiple downstream ECG datasets. Ablation studies further demonstrate the necessity and complementarity of each component. This approach provides a robust and physiologically meaningful representation learning framework for ECG analysis.
\end{abstract}


% \begin{highlights}
% \end{highlights}

%% Keywords
\begin{keyword}


Self-supervised learning \sep Electrocardiogram \sep Data augmentation \sep Mask \sep Representation learning
\end{keyword}

\end{frontmatter}



\section{Introduction}
Electrocardiogram (ECG) is a widely used non-invasive tool for cardiac disease diagnosis \cite{review1}. 
Despite recent advances in automated ECG interpretation, the development of robust data-driven models remains fundamentally constrained in practice. ECG signals are sensitive medical data subject to strict privacy regulations, which limits the availability of large-scale public datasets. Moreover, ECG annotation requires specialized expertise from cardiologists, making the labeling process costly and time-consuming \cite{review2, review3}. 
As a result, severe label scarcity persists, hindering the generalization and stability of supervised learning approaches.

Supervised deep learning methods have achieved promising performance in ECG analysis when sufficient labeled data are available \cite{supervised1, supervised2, supervised3, liu2025multimodal}. However, their effectiveness strongly depends on the scale of annotated datasets, and model performance often degrades under limited supervision. Such data scarcity is common in clinical practice, which motivates the development of learning paradigms that can effectively exploit large amounts of unlabeled ECG data.

Self-supervised learning (SSL) has emerged as a powerful framework for representation learning without manual annotations, demonstrating strong transferability across domains such as natural language processing and computer vision \cite{bert, simclr, mae, clip}. 
By designing appropriate pretraining objectives, SSL enables models to capture intrinsic structural patterns from raw data, making it particularly appealing for ECG analysis, where labeled data are scarce but unlabeled recordings are abundant.

Existing ECG SSL methods can be broadly categorized into two paradigms: contrastive learning and reconstructive learning \cite{al2025benchmarking}. 
Contrastive learning focuses on instance-level discrimination by aligning representations of augmented views and separating different samples, which encourages the learning of globally discriminative features. 
Reconstructive learning relies on masked reconstruction objectives, requiring the model to recover missing signal segments and capture local waveform structures. 
These two paradigms describe ECG representations from complementary perspectives, namely global semantic information and local structural patterns. 
However, most existing approaches adopt only a single pretraining objective, which limits the diversity of supervisory signals during pretraining \cite{jamc}. 
This observation naturally raises an important question: \textbf{Can contrastive learning and reconstructive learning be unified within a single self-supervised framework to jointly exploit their complementary strengths?}

Moreover, directly adopting generic self-supervised strategies designed for general time-series data is suboptimal for ECG signals. 
In contrastive learning, inappropriate data augmentations may disrupt the intrinsic rhythmic structure of cardiac cycles and distort clinically meaningful waveform patterns \cite{chen2025temporal}. 
In reconstructive learning, naive random masking strategies can introduce shortcut solutions in multi-lead ECG modeling due to strong inter-lead correlations, preventing the model from learning semantically meaningful representations \cite{qiu2025enhancing}.

\begin{figure}[!t]
\centering
    {\includegraphics[width=1\columnwidth]{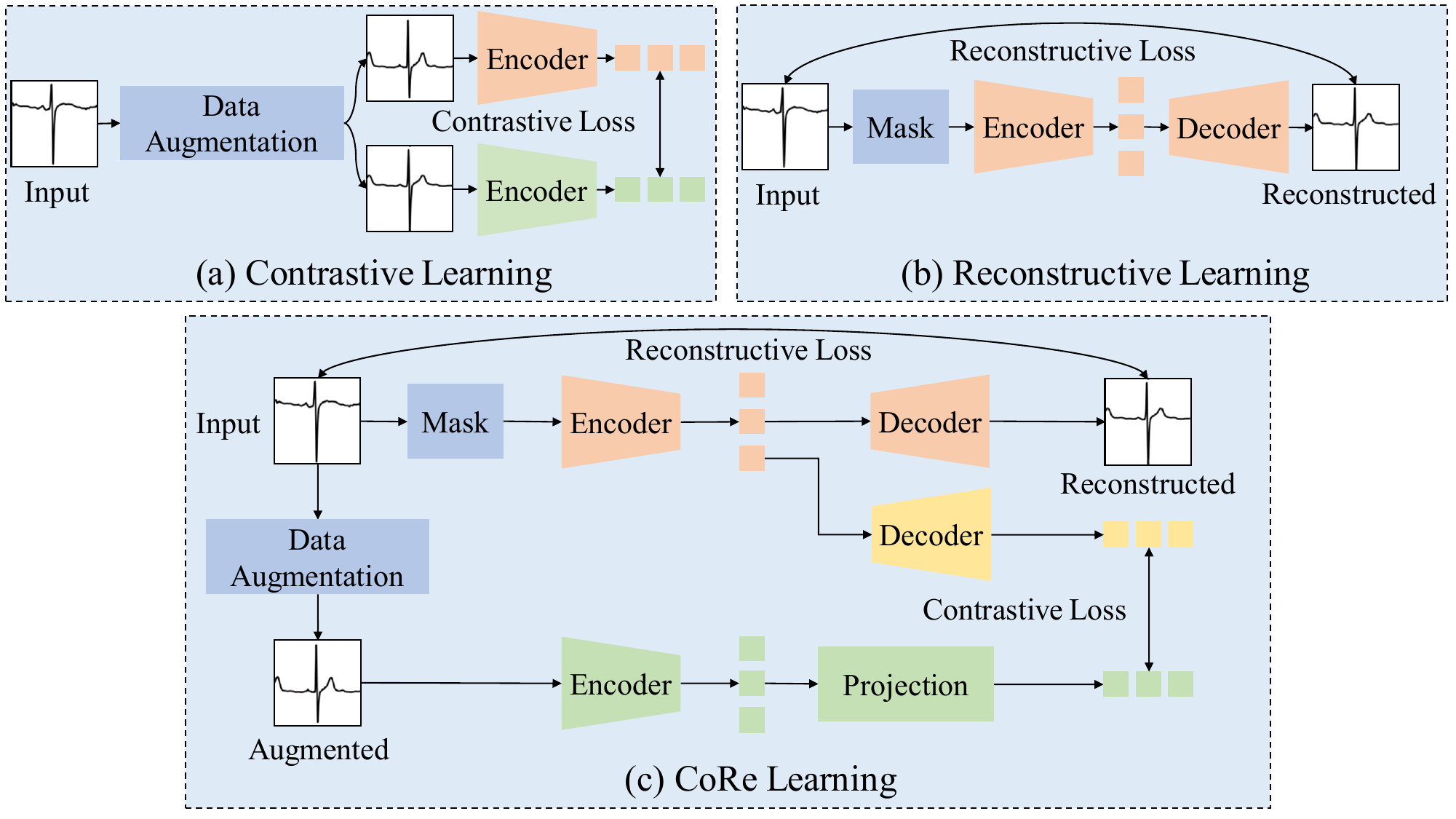}}
    \caption{Overview of CoRe Learning. Our approach integrates Contrastive Learning and Reconstructive Learning in a unified framework. The model reconstructs the original ECG signal while simultaneously aligning the global representations of the original signal and its augmented view. This design allows the two learning paradigms to remain compatible and to mutually reinforce each other.}
    \label{fig:fig_1}
\end{figure}

To address these challenges, we propose \textbf{CoRe-ECG}, a unified Contrastive and Reconstructive SSL pretraining framework for ECG representation learning. As shown in Fig.~\ref{fig:fig_1}(c), CoRe-ECG tightly integrates contrastive learning and reconstructive learning, allowing instance-level discriminative signals to directly guide the reconstruction process and enabling cooperative learning of global semantic representations and local waveform structures. 
To further enhance physiological consistency and representation robustness, we introduce a frequency dynamic augmentation mechanism and a spatio-temporal dual masking strategy, which mitigate non-physiological distortions and shortcut learning in multi-lead ECG modeling. 
The entire framework is trained end-to-end with unified contrastive and reconstructive objectives, yielding expressive and transferable ECG representations for downstream classification tasks. We conduct systematic experiments on multiple public ECG datasets. The results demonstrate that CoRe-ECG achieves state-of-the-art performance on downstream ECG classification tasks and provide strong evidence for the necessity and complementary effects of the proposed components.

Our main contributions are summarized as follows.

\begin{enumerate}[(1)]
% \item We propose a flexible guidance module to generate graspable areas for different grasp guidance modalities such as human instructions, interactive clicks, and parts affordance.
\item We propose CoRe-ECG, a unified contrastive and reconstructive self-supervised pretraining paradigm that injects instance-level discriminative signals from contrastive learning directly into ECG reconstruction, enabling cooperative learning of global semantics and local waveform structures.
% In this way, CoRe-ECG establishes a cooperative relationship between global semantic modeling and local structural learning.
\item We design a frequency dynamic augmentation mechanism that explicitly models the importance distribution of ECG in frequency domain and applies adaptive perturbations, effectively enhancing representation robustness while avoiding non-physiological waveform distortions.
\item We design a spatio-temporal dual masking strategy along both lead and temporal dimensions, which breaks trivial linear dependencies and forces the model to learn cross-lead and cross-temporal reconstruction.
\end{enumerate}

% \begin{enumerate}
% \itemsep=0pt
% \item {natbib.sty} for citation processing;
% \item {geometry.sty} for margin settings;
% \item {fleqn.clo} for left aligned equations;
% \item {graphicx.sty} for graphics inclusion;
% \item {hyperref.sty} optional packages if hyperlinking is
%   required in the document;
% \end{enumerate}

% The Elsevier cas-dc class is based on the
% standard article class and supports almost all of the functionality of
% that class. In addition, it features commands and options to format the
% \begin{itemize} \item document style \item baselineskip \item front
% matter \item keywords and MSC codes \item theorems, definitions and
% proofs \item lables of enumerations \item citation style and labeling.
% \end{itemize}

% This class depends on the following packages
% for its proper functioning:

% \begin{enumerate}
% \itemsep=0pt
% \item {natbib.sty} for citation processing;
% \item {geometry.sty} for margin settings;
% \item {fleqn.clo} for left aligned equations;
% \item {graphicx.sty} for graphics inclusion;
% \item {hyperref.sty} optional packages if hyperlinking is
%   required in the document;
% \end{enumerate}  

% All the above packages are part of any
% standard \LaTeX{} installation.
% Therefore, the users need not be
% bothered about downloading any extra packages.

\section{Related work}

\subsection{ECG Contrastive Learning}

Contrastive learning aims to learn discriminative representations by minimizing the distance between similar samples while maximizing the distance between dissimilar ones. In practice, this paradigm relies on the construction of positive and negative sample pairs. For ECG representation learning, positive pairs are typically derived from the same underlying signal segment, whereas negative pairs are sampled from different signal instances \cite{ecgfm}. To generate positive pairs, prior works commonly apply diverse transformations to a single ECG recording to produce multiple views, such as data augmentation. Typical augmentation strategies include random cropping, additive Gaussian noise, and temporal scaling through stretching or compression along the time axis \cite{contrast_everything}. By contrasting these augmented views, models are encouraged to learn global representations that are robust to noise and minor morphological variations.

MoCo v3 \cite{moco} introduces a momentum-updated target encoder together with a predictor network to stabilize contrastive training. Multiple augmented views of the same ECG signal are generated, and the model is trained to align representations of the original signal and its augmented counterparts under a self-supervised contrastive learning framework. In contrast, CMSC \cite{cmsc} avoids explicit data augmentation and instead exploits the temporal structure of ECG signals. It segments a single ECG recording into multiple non-overlapping temporal subsegments and constructs positive pairs using representations of subsegments from the same patient, thereby capturing time-invariant characteristics of ECG signals. Several methods extend contrastive learning to multi-lead ECG signals. LFBT \cite{lfbt} proposes a lead-fusion contrastive framework that integrates intra-lead and inter-lead Barlow Twins losses, jointly enforcing representation consistency within individual leads and alignment across different leads. This design enables the encoder to capture both lead-specific patterns and cross-lead correlations. For downstream tasks, LFBT adopts a multi-branch concatenation (MBC) strategy, where each lead is processed by an independent encoder and the resulting representations are concatenated to fuse information from all leads. Zhang et al. \cite{zhang} proposes a self-supervised ECG representation learning method based on manipulated temporal-spatio reverse detection. By applying temporal reversal, spatial reversal, and combined temporal-spatio reversal to ECG signals, the model is trained to distinguish original signals from their manipulated versions, formulating representation learning as a pretext classification task. MRC \cite{mrc} focuses on jointly modeling morphological and rhythmic characteristics of multi-lead ECG signals. It employs random beat selection and 0-1 pulse generation to create augmented views, and adopts a three-branch architecture to map different views into a shared latent space, enabling dual contrastive learning across morphology and rhythm.

Despite the effectiveness of these contrastive learning approaches, their performance heavily depends on the construction of positive sample pairs. Directly leveraging existing data to form positive and negative pairs may limit the diversity of variations observed during training, preventing models from capturing richer semantic information. Moreover, overly aggressive augmentation strategies, such as spatio-temporal reversal or the addition of synthetic sine waves, may introduce physiologically implausible positive pairs and lead the model toward suboptimal representations. Consequently, designing physiologically meaningful data augmentation strategies that provide sufficient and informative variations remains a central challenge for contrastive learning in ECG representation learning. 

\subsection{ECG Reconstructive Learning}

Reconstructive learning represents another paradigm of self-supervised learning, where the core idea is to recover artificially masked or corrupted data in order to capture latent structural characteristics. In practice, this paradigm typically applies masking or perturbation operations to the original input, generating visible and masked regions. The model is trained to extract representations from the visible parts and reconstruct the masked portions, thereby encouraging the learning of fine-grained local representations and contextual dependencies within the input data.

Oh et al. \cite{oh} propose a lead-agnostic self-supervised framework that randomly masks entire ECG signals from different leads and leverages a hybrid CNN-Transformer architecture to jointly learn local and global representations. MaeFE \cite{maefe} applies masked autoencoding to ECG self-supervised learning by performing reconstruction tasks through temporal or spatial masking. However, this approach does not explicitly model the joint interaction between temporal and spatial dimensions. ECG-MAE \cite{ecg-mae} extends this paradigm by simultaneously applying random masking along both the temporal and lead dimensions, enabling the model to capture spatiotemporal information in multi-lead ECG signals. Additionally, ECG-MAE investigates the performance differences between random masking and structured grid masking with continuous masked regions. ST-MEM \cite{st-mem} introduces a lead indicator module to explicitly model both intra-lead consistency and inter-lead variability. By sharing a reconstruction decoder across the same leads, this method enhances the reconstruction process and improves the utilization of lead-specific and cross-lead information. BMIRC \cite{bmirc} further enriches reconstructive learning by incorporating frequency-domain representations obtained via Fourier transformation into the masked reconstruction task. Moreover, it establishes internal representation connections (IRC) between the encoder and decoder to facilitate information flow and improve reconstruction quality. HeartLang \cite{heartlang} formulates ECG representation learning from a language modeling perspective. It segments ECG signals by identifying QRS complexes and introduces a quantized codebook to represent ECG patches as discrete tokens. Reconstruction is then performed at the token level, encouraging the model to learn higher level semantic representations of ECG signals.

Despite their promising performance, reconstructive methods critically depend on the design of masking strategies. Most existing approaches adopt random masking schemes, which may be suboptimal for ECG signals characterized by strong regularity and periodic structure. Such strategies often fail to account for the potential linear coupling among different leads, making the reconstruction task overly trivial. As a result, models may rely on simple signal regularities to complete reconstruction, rather than learning essential physiological patterns. In masked image modeling, recent studies have begun to explore how masking strategies influence self-supervised representation learning \cite{qiu2024masked, mask}. Similarly, for ECG signals, disrupting simplistic correlations between visible and masked regions and forcing models to capture long range temporal dependencies and cross-lead contextual information has become a key research focus.

% The author names and affiliations could be formatted in two ways:
% \begin{enumerate}[(1)]
% \item Group the authors per affiliation.
% \item Use footnotes to indicate the affiliations.
% \end{enumerate}
% See the front matter of this document for examples. 
% You are recommended to conform your choice to the journal you 
% are submitting to.

\section{Method}

\subsection{The architecture of CoRe-ECG}

\begin{figure}[!t]
\centering
    {\includegraphics[width=1\columnwidth]{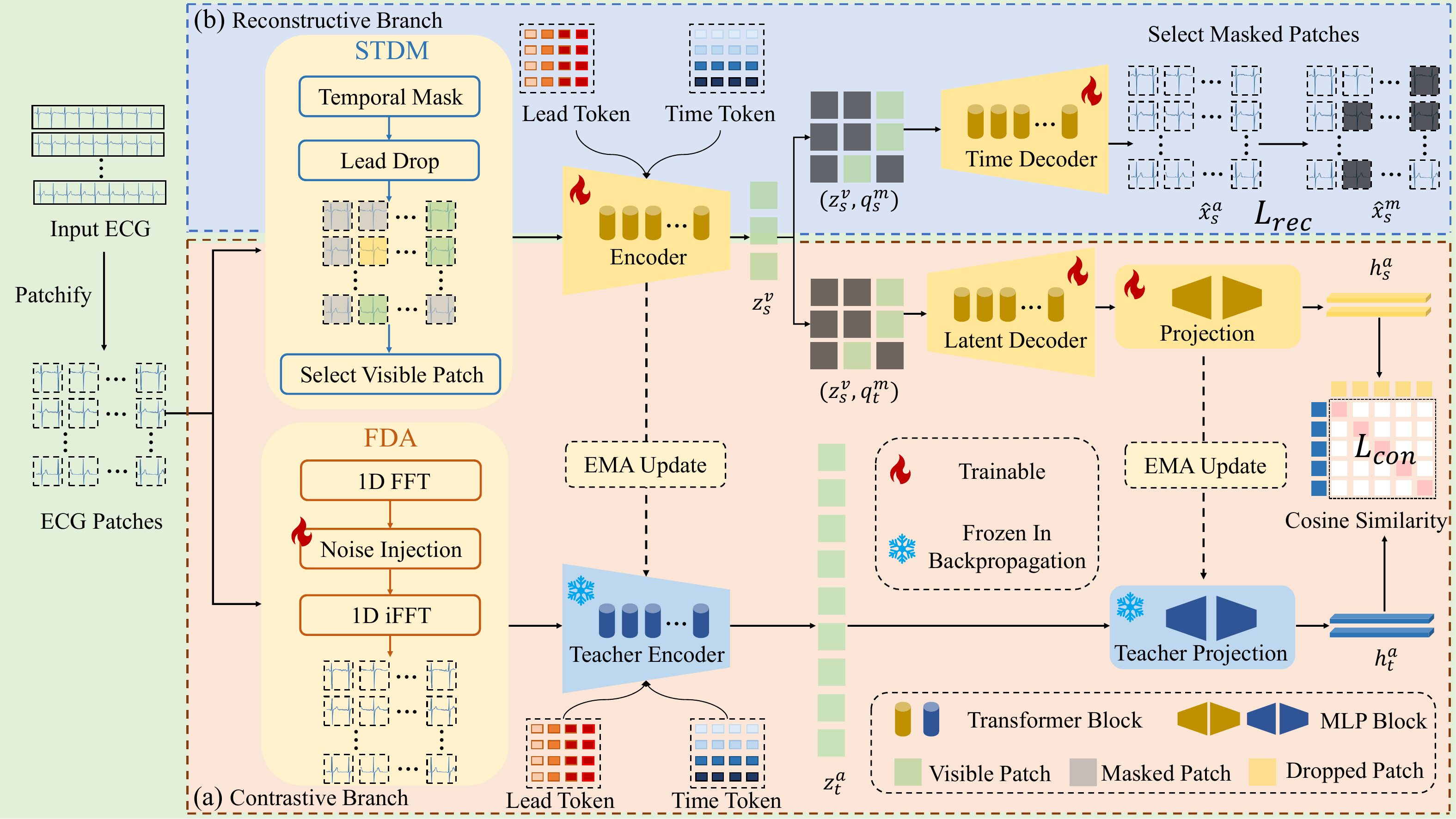}}
    \caption{Overview of the CoRe-ECG architecture, which consists of two parallel branches: the Contrastive Branch and the Reconstructive Branch. In (a), we utilize frequency dynamic augmentation to generate an augmented view, which is transformed into a global representation via the encoder-projection structure. Simultaneously, the visible representations of the raw signal extracted by the Encoder are passed through the decoder-projection module to yield a corresponding global representation. The contrastive loss is calculated between these two representations. In (b), the input undergoes spatio-temporal dual masking and is processed by the encoder-decoder network to reconstruct the original input signal. The reconstruction loss is computed exclusively on the masked patches. }
    \label{fig:overview of architecture}
\end{figure}

As shown in Fig.~\ref{fig:overview of architecture}, the proposed CoRe-ECG framework integrates two complementary learning branches: the Contrastive Branch and the Reconstructive Branch, enabling robust ECG representation learning during pre-training.

In the Contrastive Branch, we employ frequency dynamic augmentation to generate the augmented view from the raw view. To produce a global, high-semantic feature representation that serves as a contrastive target for the original view, the model is compelled to learn instance-discriminative features.  Since the Encoder only observes visible patches from the raw ECG due to masking, the Teacher Encoder processes the full augmented signal to provide a complete semantic target representation, ensuring that the student learns to infer global context from partial observations. Subsequently, we utilize the Latent Decoder to complement the output features of the Encoder before computing the contrastive loss against the global features of the augmented view.

In the Reconstructive Branch, starting from the raw 12-lead ECG, we first apply a Spatio-Temporal Dual Masking policy that partitions patches into visible, masked, and dropped patches. Visible patches are encoded by the Encoder to obtain visible ECG representations, which are processed by the Time Decoder to reconstruct the raw ECG. The reconstruction loss is computed exclusively on the masked patches, preventing linear dependencies across leads from leaking and thereby compelling the model to learn comprehensive cross-lead and cross-temporal ECG representations.

In detail, in the Reconstructive Branch, given an input ECG signal $\mathbf{X} \in \mathbb{R}^{C \times T}$, where $C$ denotes the number of ECG leads and $T$ represents the temporal length of the signal, we first partition it into patches. Specifically, a non-overlapping temporal segmentation is adopted to divide the signal into temporal patches $x_s^a \in \mathbb{R}^{C \times N \times P}$, where $N$ denotes the number of patches, $P$ denotes the temporal length of each patch, and the superscript $a$ indicates all patches.

Most reconstructive learning frameworks typically adopt random masking strategies, applying random masking with equal probability along both the temporal and lead dimensions simultaneously. However, in multi-lead ECG modeling, such approaches suffer from a linear dependence issue. Due to the physical derivation of ECG leads (e.g., Einthoven's law \cite{einthoven_law}, where Lead I + Lead III = Lead II), a model may trivially infer masked regions through simple linear combinations of visible leads, without learning robust semantic representations. Spatio-Temporal Dual Masking (which will be detailed in Section~\ref{sec:stdm}) is applied to all ECG patches $x_s^a$, resulting in visible ECG patches $x_s^v$, masked ECG patches $x_s^m$, and dropped ECG patches $x_s^d$. The visible patches $x_s^v$ are then fed into the Encoder $E_s$ to extract ECG representations $z_s^v$.
\begin{equation}
z_s^v = E_s(x_s^v; \theta_{E_s}),
\end{equation}
The Encoder $E_s$ consists of a patch embedding layer implemented using 1D convolutional layers, followed by multiple Transformer layers, aiming to capture the global contextual information of the ECG signal from visible patches.

Subsequently, the representations $z_s^v$ are concatenated with randomly initialized mask embeddings $q_s^m$ corresponding to the masked and dropped positions. The concatenated embeddings are then passed to the Time Decoder $D_t$ to reconstruct the raw ECG signal $\hat{x}_s^a$.
\begin{equation}
\hat{x}_s^a = D_t(z_s^v, q_s^m; \theta_{D_t}),
\end{equation}

In the Contrastive Branch, standard contrastive frameworks process two augmented views symmetrically. Time-series data augmentation methods commonly adopt heuristic geometric transformations, including jittering, scaling, and permutation. Although these techniques have shown effectiveness in general time-series modeling, they are less suitable for ECG analysis. In particular, permutation-based operations can alter the intrinsic temporal structure of the P-QRS-T complex, compromising clinically meaningful patterns. In addition, simple noise injection can alter the subtle morphological characteristics of ECG signals, resulting in a distribution mismatch that is inconsistent with the physiological signals generation mechanisms. To avoid semantic damage to the ECG caused by traditional data augmentation methods, we utilize Frequency Dynamic Augmentation (which will be detailed in Section~\ref{sec:fda}) to generate an augmented view $x_t^a$ of the raw ECG $x_s^a$. Here, we do not adopt any masking method, in order to obtain the global representation of the augmented view as the learning target for the raw ECG representation. We pass $x_t^a$ through the Teacher Encoder $E_t$ to obtain the ECG representation $z_t^a$. 
\begin{equation}
z_t^a = E_t(x_t^a; \theta_{E_t}),
\end{equation}
To prevent representation collapse and stabilize training, we adopt a momentum update mechanism. The parameters of the Teacher Encoder $\theta_{E_t}$ are not updated via back-propagation but are an exponential moving average (EMA) \cite{byol} of the Encoder parameters $\theta_{E_s}$. 
\begin{equation}
    \theta_{E_t} \leftarrow m \cdot \theta_{E_t} + (1 - m) \cdot \theta_{E_s},
\end{equation}
Simultaneously, we concatenate the visible ECG representations $z_s^v$ obtained above with randomly initialized mask embeddings $q_t^m$. We feed the concatenated embeddings into the Latent Decoder $D_l$ and use mean pooling to reconstruct the global ECG embedding $\hat{z}_s^a$. 
\begin{equation}
\hat{z}_s^a = D_l(z_s^v, q_t^m; \theta_{D_l}),
\end{equation}
This asymmetry forces the model to not only reconstruct the signal but also infer a global semantic representation from partial observations that matches the robust features extracted from the frequency-augmented signal. To align the representations under different views, we map $\hat{z}_s^a$ and $z_t^a$ to the same space via the Projection $H_s$ and the Teacher Projection $H_t$ respectively, obtaining $H_s^a$ and $H_t^a$. 
\begin{equation}
H_s^a = H_s(\hat{z}_s^a; \theta_{H_s}),
\end{equation}
\begin{equation}
H_t^a = H_t(z_t^a; \theta_{H_t}),
\end{equation}
Similarly, the parameters of the Teacher Projection are updated via EMA from the Projection.
\begin{equation}
    \theta_{H_t} \leftarrow m \cdot \theta_{H_t} + (1 - m) \cdot \theta_{H_s},
\end{equation}

Through the above process, the Time Decoder produces the reconstructed ECG patches $\hat{x}_s^a$, which are used to compute the reconstruction loss over the masked positions. Meanwhile, the Projection Head and Teacher Projection Head generate global representations $H_s^a$ and $H_t^a$ under the raw and frequency-augmented views, respectively. These representations serve as the targets for contrastive learning, enabling the model to jointly learn accurate signal reconstruction and robust global semantic representations.

\subsection{Spatio-Temporal Dual Masking (STDM)}
\label{sec:stdm}

To prevent trivial inference caused by inter-lead linear relationships, we propose Spatio-Temporal Dual Masking (STDM). Unlike uniform random masking, STDM dynamically alternates between full temporal mask, which forces the model to rely on long-range temporal context, and partial mask with lead drop, which encourages the learning of non-linear spatial dependencies across leads.

Formally, given an input ECG signal $\mathbf{X} \in \mathbb{R}^{C \times T}$, we partition it into non-overlapping temporal patches $x_s^a \in \mathbb{R}^{C \times N \times P}$. For each patch index $n \in \{1, \dots, N\}$, we construct a visible index set $V_n$ as input to the Encoder and a masked index set $M_n$ for reconstruction supervision through the following hierarchical procedure:

\begin{figure}[t]
\centering
{\includegraphics[width=1.0\columnwidth]{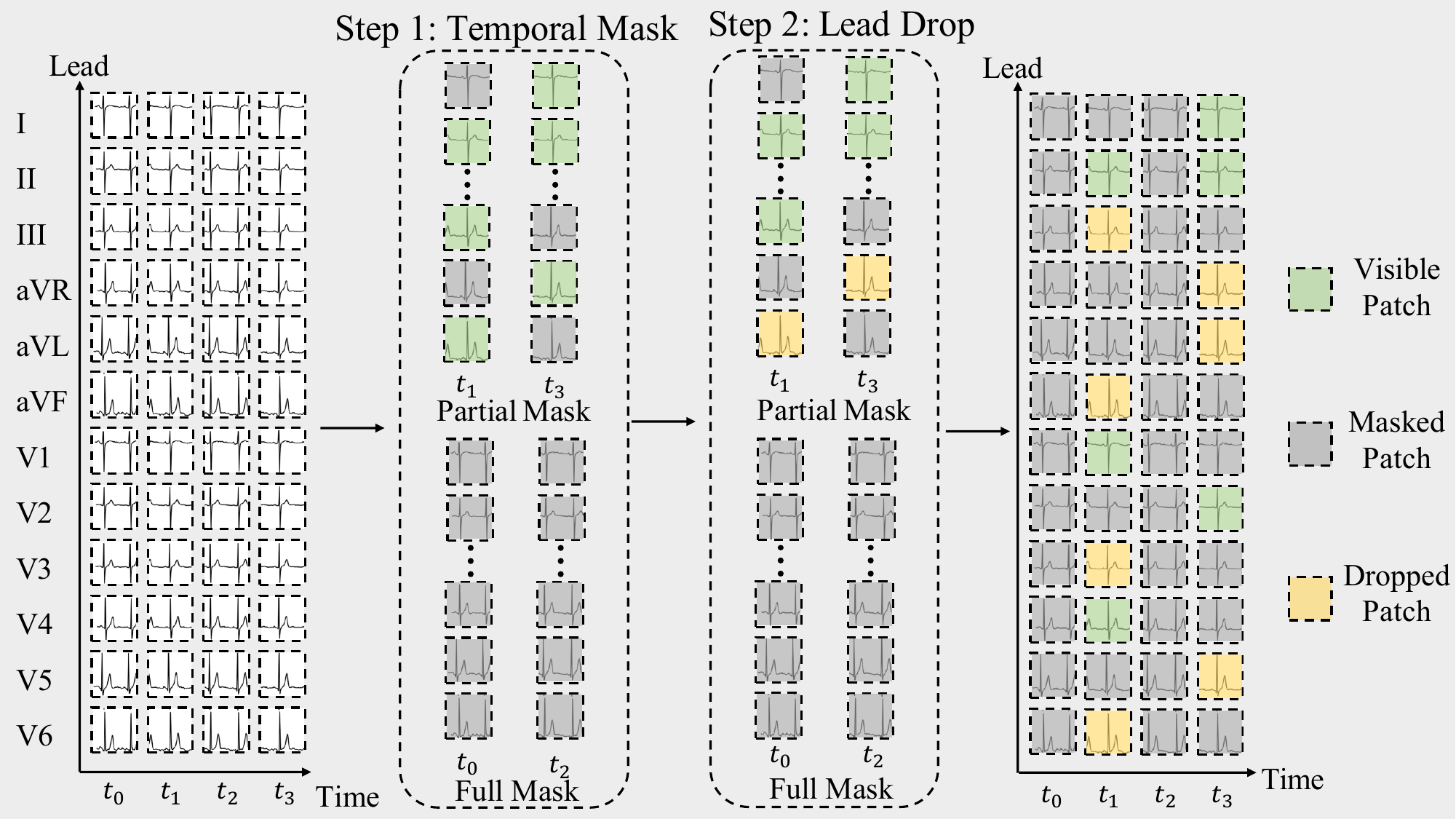}}
\caption{The structure of Spatio-Temporal Dual Masking (STDM).}
\label{fig:stdm}
\end{figure}

\textbf{Step 1 Temporal Mask}: For each temporal index $n$, we sample a mode indicator $\delta_n \sim \text{Bernoulli}(P_{\text{time}})$. If $\delta_n = 1$ (full temporal mask), the entire time step is masked, yielding an empty visible set $V_n = \emptyset$, while all leads are assigned to the reconstruction target $M_n = \{1, \dots, C\}$.

\textbf{Step 2 Lead Drop}: If $\delta_n = 0$ (partial mask), we randomly sample a subset of $k$ leads to remain visible, denoted as $V_n \subset \{1, \dots, C\}$ with $|V_n| = k$. The remaining leads $R_n = \{1, \dots, C\} \setminus V_n$ constitute candidates for reconstruction. To prevent the model from exploiting trivial spatial correlations among the masked leads, we do not reconstruct the entire candidate set $R_n$. Instead, we apply a lead drop strategy:
\begin{equation}
D_n = \{ c \in R_n \mid z_c = 1, z_c \sim \text{Bernoulli}(P_{\text{lead}}) \},
\end{equation}
where $D_n$ denotes the set of dropped indices and $M_n = R_n \setminus D_n$ denotes the set of masked indices. By stochastically excluding easy-to-infer leads from the supervision signal, STDM compels the Encoder to capture morphological characteristics rather than relying on linear combinations.

Finally, we aggregate the indices from the above sets and construct the visible matrix $V = \{V_n\} \in \{0, 1\}^{C \times N}$, masked matrix $M = \{M_n\} \in \{0, 1\}^{C \times N}$, and dropped matrix $D = \{D_n\} \in \{0, 1\}^{C \times N}$. The visible, masked, and dropped ECG patches are obtained by an element-wise masking operation:
\begin{equation}
x_s^v = x_s^a \odot V, \quad x_s^m = x_s^a \odot M, \quad x_s^d = x_s^a \odot D,
\end{equation}
where $\odot$ denotes element-wise multiplication.

\subsection{Frequency Dynamic Augmentation (FDA)}
\label{sec:fda}

\begin{figure}[!t]
  \centering
  \includegraphics[width=1.0\columnwidth]{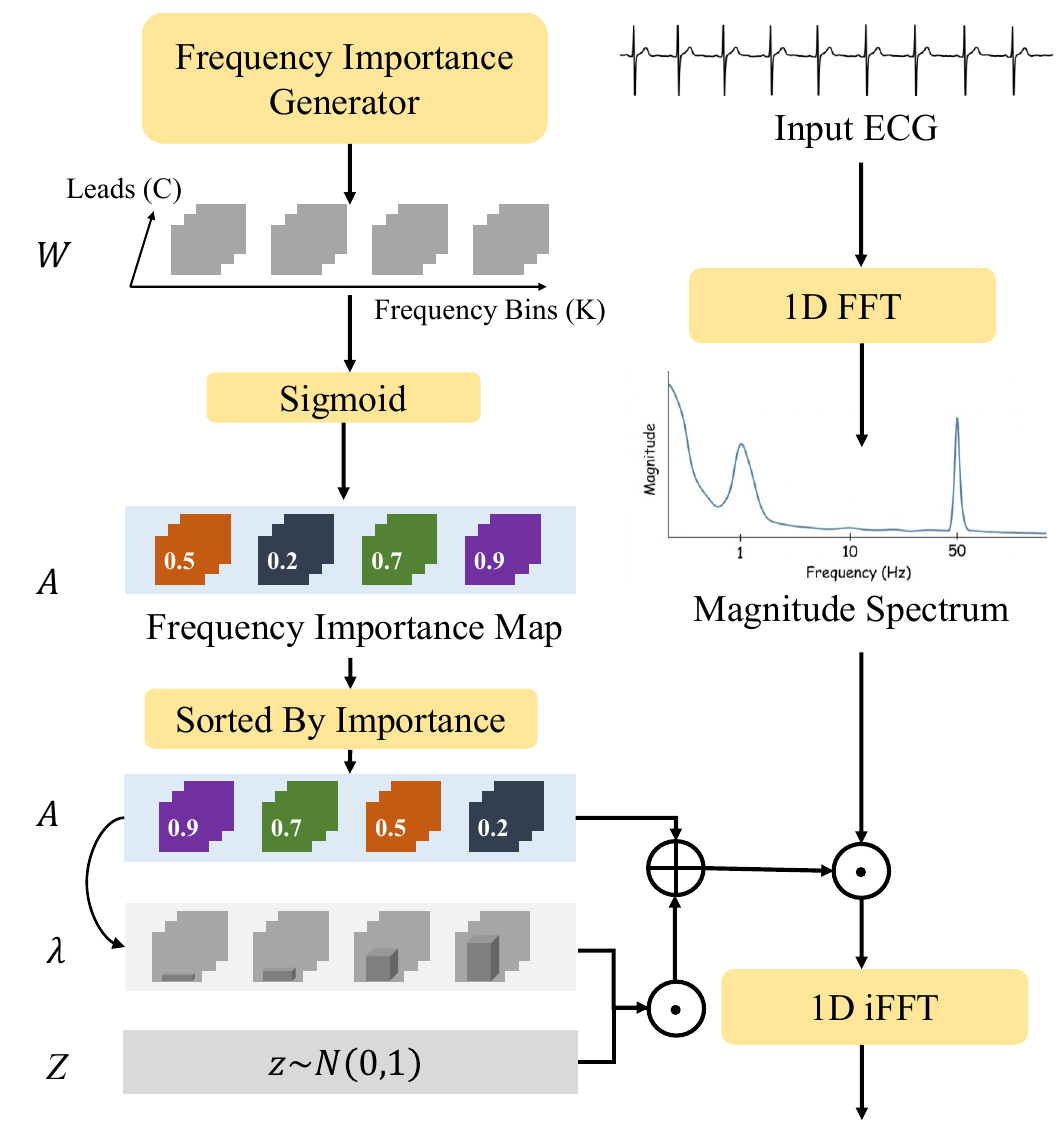}
  \caption{Illustration of Frequency Dynamic Augmentation (FDA). By modulating frequency components in the magnitude spectrum based on learnable importance, FDA preserves critical physiological semantics while introducing diverse perturbations.}
  \label{fig:fda}
\end{figure}

To preserve the physiological characteristics of ECG signals while enabling controllable perturbations, we design Frequency Dynamic Augmentation (FDA), a method that performs perturbations in the frequency domain with frequency components weighted by their relative importance. This approach is motivated by the observation that ECG signals exhibit high energy concentration in specific frequency bands (typically 0.65-40 Hz). Transforming signals into the frequency domain allows us to decouple global semantic information from local noise, enabling more controllable augmentation.

Formally, given the input ECG signal $\mathbf{X}$, we first compute its frequency representation $X^f \in \mathbb{C}^{C \times K}$ via the Fast Fourier Transform (FFT), where $K = \lfloor T / 2 \rfloor + 1$. To distinguish between informative signals and noise components, we generate a global learnable weight matrix $W \in \mathbb{R}^{C \times K}$. The frequency importance map $A \in [0, 1]^{C \times K}$ is obtained via a sigmoid activation $A = \sigma(W)$. A higher value in $A$ indicates that the corresponding frequency component encodes critical morphological information.

Unlike standard dropout or additive noise, FDA employs an inverse-weighted noise injection strategy. We consider that critical frequencies should remain stable, while less informative frequencies can tolerate higher variance. We define a frequency-wise noise scaling factor $\lambda\in \mathbb{R}^{C \times K}$, which is inversely proportional to the importance map A:
\begin{equation}
\lambda = \frac{1}{\epsilon + A},
\end{equation}
where $\epsilon$ is a stability term. To further protect critical frequency components from being perturbed, 
we take the 50th percentile of $A$ across frequency components in each lead as the threshold $\theta$ to distinguish important from non-important frequencies.
Specifically, for each frequency 
component, if its importance weight exceeds $\theta$, the corresponding 
noise scaling factor is set to zero:
\begin{equation}
\lambda =
\begin{cases}
0, & A \ge \theta, \\
(\epsilon + A)^{-1}, & \text{otherwise}.
\end{cases}
\end{equation}

Moreover, to prevent excessive noise injection, the non-zero elements 
of $\lambda$ are normalized within each lead to ensure 
a stable perturbation magnitude.

\begin{algorithm}[!htbp]
\small
\setlength{\baselineskip}{11pt}
\caption{Frequency Dynamic Augmentation (FDA)}
\label{alg:fda}
\small
\begin{algorithmic}[1]

\STATE \textbf{Input:} ECG $\mathbf{X} \in \mathbb{R}^{C \times T}$,
frequency importance $\mathbf{A} \in \mathbb{R}^{C \times K}$
\STATE \textbf{Output:} Augmented ECG Patches $\mathbf{x}_t^a \in \mathbb{R}^{C \times N \times P}$

\vspace{0.4em}
\STATE $\mathbf{X}^f \leftarrow \mathrm{FFT}(\mathbf{X})$
\hspace{0.8em}\textit{// Fast Fourier Transform}

\vspace{0.4em}
\FOR{$c = 1$ to $C$}

    \STATE $\theta_c \leftarrow \mathrm{Quantile}_{0.5}(\mathbf{A}_{c,:})$
    \hspace{0.8em}\textit{// as the threshold for frequency selection}

    \STATE $\mathbf{Z}_{c,:} \sim \mathcal{N}(\mathbf{0}, \mathbf{I})$
    \hspace{0.8em}\textit{// Gaussian noise}

    \FOR{$k = 1$ to $K$}

        \IF{$\mathbf{A}_{c,k} \ge \theta_c$}
            \STATE $\boldsymbol{\lambda}_{c,k} \leftarrow 0$
            \hspace{0.8em}\textit{// do not perturb high-importance frequency}
        \ELSE
            \STATE $\boldsymbol{\lambda}_{c,k} \leftarrow 
            (\mathbf{A}_{c,k} + \epsilon)^{-1}$
            \hspace{0.8em}\textit{// inverse-proportional noise strength}
        \ENDIF

    \ENDFOR

    \STATE $\mu_c \leftarrow 
    \mathrm{mean}(\boldsymbol{\lambda}_{c,k} \mid \boldsymbol{\lambda}_{c,k} \neq 0)$
    \hspace{0.8em}\textit{// mean over non-zero noise coefficients}

    \FOR{$k = 1$ to $K$}
        \IF{$\boldsymbol{\lambda}_{c,k} \neq 0$}
            \STATE $\boldsymbol{\lambda}_{c,k} \leftarrow 
            \boldsymbol{\lambda}_{c,k} / \mu_c$
            \hspace{0.8em}\textit{// normalize non-zero frequency perturbation}
        \ENDIF
    \ENDFOR

    \STATE $\mathbf{X}^f_{aug} \leftarrow
    \mathbf{X}^f_{c,:} \odot
    \left(
    \mathbf{A}_{c,:}
    +
    \boldsymbol{\lambda}_{c,:} \odot \mathbf{Z}_{c,:}
    \right)$
    \hspace{0.8em}\textit{// frequency-domain perturbation}

\ENDFOR

\vspace{0.4em}
\STATE $\mathbf{x}_t^a \leftarrow 
\mathrm{Patchify}(\mathrm{iFFT}(\mathbf{X}^f_{aug}))$

\end{algorithmic}
\end{algorithm}
The final augmented frequency is obtained by modulating the original frequency with the fused importance-noise map:
\begin{equation}
X^f_{aug} = X^f \odot (A + \lambda \odot Z),
\end{equation}
where $\odot$ denotes element-wise multiplication and $Z \sim \mathcal{N}(0, 1)$ represents randomly sampled Gaussian noise. Here, $A$ represents the importance of frequency components, while $\lambda \odot Z$ represents the intensity of the injected noise. For important frequency components, $A$ is close to 1 and $\lambda$ is set to 0, ensuring that the original morphology is preserved. Conversely, for unimportant frequencies, as $A$ approaches 0, $\lambda$ becomes larger, resulting in greater morphological variation. Finally, the augmented ECG patches $x_t^a$ is obtained via the Inverse FFT (iFFT) of $X^f_{aug}$. By end-to-end training, FDA learns to preserve diagnostically important frequencies while diversifying the background noise profile, thereby improving model robustness.

\subsection{Training and Inference}

Based on the above two branches, we obtain the reconstruction result of the original ECG, denoted as $\hat{x}_s^a$. As described in Section~\ref{sec:stdm}, to prevent the model from exploiting linear correlations between leads to cheat, we compute the reconstruction loss solely on the masked patches $x_s^m$. This is formulated as:

\[
L_{rec} = \frac{1}{|M|} \left\| (x_s^a - \hat{x}_s^a) \odot M \right\|_F^2,
\]

Additionally, we calculate the contrastive loss based on $H_s^a$ and $H_t^a$ for all samples in the batch. Specifically, we optimize the InfoNCE loss to maximize the similarity between each $H_s^a$ and its corresponding positive target $H_t^a$, which is derived from the same ECG signal via data augmentation, while minimizing the similarity with negative samples from other ECG signals within the current batch. The batch-level loss is defined as:

\begin{equation}
\label{eq:contrastive_loss}
L_{con} = \frac{1}{|\mathcal{B}|} \sum_{H_s^a \in \mathcal{B}}
- \log 
\frac{
\exp(\mathrm{sim}(H_s^a, H_t^a) / \tau)
}{
\sum_{H' \in \mathcal{B}} \exp(\mathrm{sim}(H_s^a, H') / \tau)
},
\end{equation}

where $\mathcal{B}$ denotes the set of all sample representations in the current batch, including both positive and negative samples and $\tau$ is the temperature coefficient controlling the sharpness of the similarity distribution.

In summary, our total loss is given by:

\begin{equation}
    L = \alpha \cdot L_{rec} + \beta \cdot L_{con},
\end{equation}
where $\alpha$ and $\beta$ are hyperparameters.

\begin{figure}[t]
  \centering
  \includegraphics[width=1.0\columnwidth]{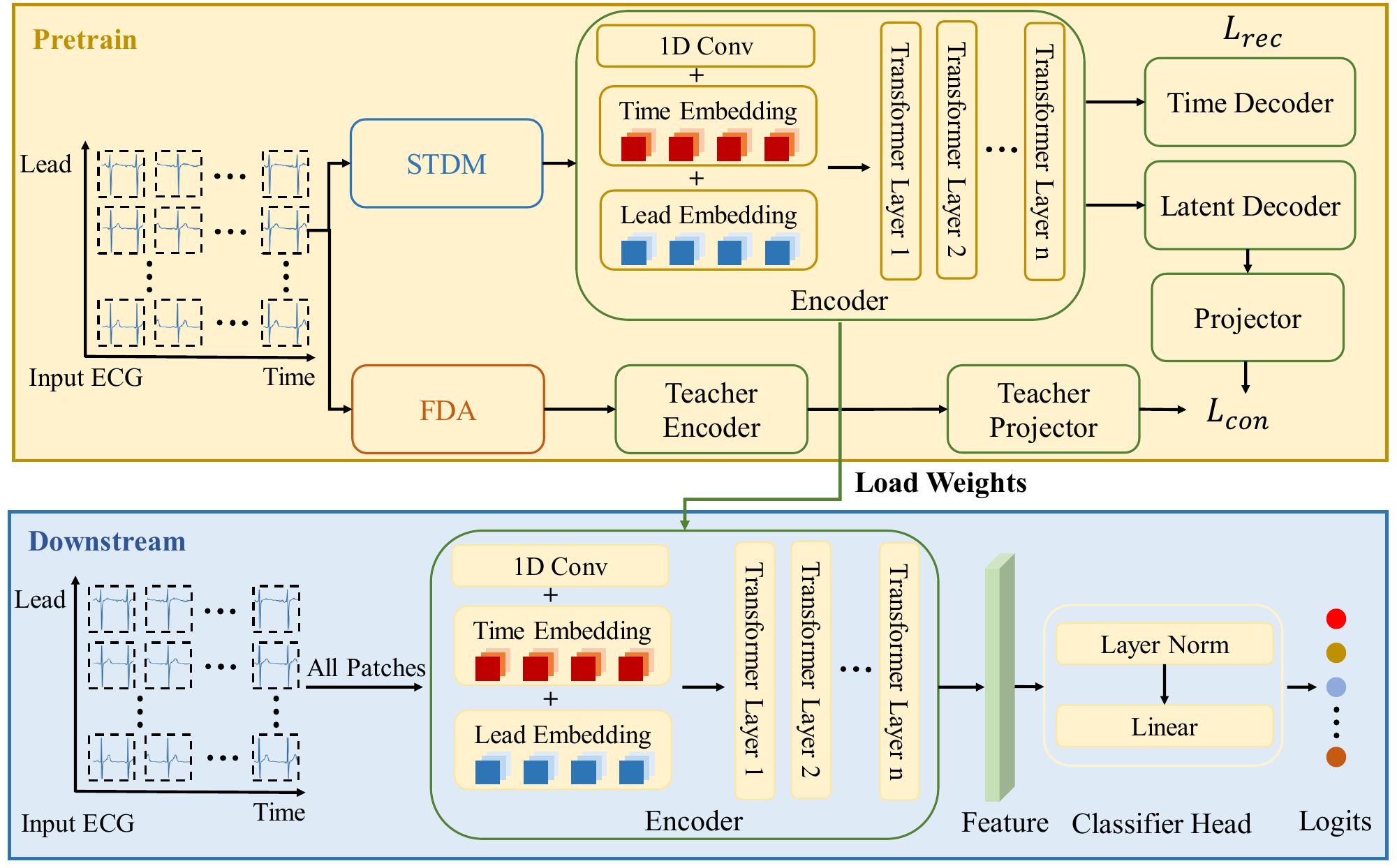}
  \caption{Schematic illustration of the downstream fine-tuning phase.}
  \label{fig:down}
\end{figure}

In the downstream phase, we employ the pre-trained Encoder coupled with a randomly initialized linear layer to address the classification task. As illustrated in Fig. \ref{fig:down}, we initialize the Encoder using the pre-trained checkpoint. All ECG patches are fed into the Encoder to extract feature representations, which are subsequently passed through the linear classification head to produce the output logits. During this process, the model undergoes full fine-tuning, where the parameters of both the Encoder and the newly added linear layer are updated.

\section{Experiment}
\label{sec:experiments}

\subsection{Datasets and Preprocessing}
\label{sec:datasets}
To evaluate the effectiveness of our model, we pretrain it on the largest publicly available ECG dataset MIMIC-IV-ECG \cite{gow2023mimic}, which contains approximately 800,000 12-lead diagnostic ECG recordings from nearly 160,000 patients. For downstream tasks, we finetune and evaluate the model on three publicly available datasets:

\begin{itemize}
    \item \textbf{PTB-XL} \cite{ptbxl}: Consists of 21,837 12-lead ECG records from 18,885 patients. Each record has a 10-second duration and was originally sampled at 500~Hz. It is categorized into 5 diagnostic classes.
    \item \textbf{ICBEB2018} \cite{chinaecg}: Contains over 6,000 12-lead ECG records with durations ranging from 6 to 60 seconds, sampled at 500~Hz. It is categorized into 9 diagnostic classes.
    \item \textbf{Ningbo} \cite{zheng2020}: Sourced from the PhysioNet/Computing in Cardiology Challenge 2021 \cite{reyna2021will}. It comprises 34,905 ECG recordings, each with a 10-second duration sampled at 500~Hz. This dataset presents a multi-label classification task including 25 categories.
\end{itemize}

To standardize the input format across these heterogeneous datasets, we adopt a unified preprocessing pipeline as follows:

Fixed length. We first extract ECG signals from the above datasets. Signals shorter than 10 seconds are excluded. For recordings longer than 10 seconds, we segment the signals using a sliding-window cropping strategy with a fixed window length of 10 seconds, iteratively extracting contiguous segments until the remaining signal length is shorter than 10 seconds. This results in standardized ECG clips of fixed duration.

Denoising. We apply a cascaded high-pass and low-pass Butterworth filtering scheme to remove noise components outside the 0.65-40 Hz frequency band, effectively suppressing baseline wander and high-frequency artifacts while preserving diagnostically relevant ECG information.

Alignment. To ensure fair comparison with baseline methods, all ECG signals are resampled to 250 Hz, yielding 2,500 timesteps per recording. Following the experimental settings of ST-MEM \cite{st-mem}, we randomly crop 2,250 timesteps as model input. Lead ordering is standardized according to the conventional sequence: I, II, III, aVR, aVL, aVF, V1-V6. In addition, Z-score normalization is applied along the temporal dimension to compensate for amplitude variations across different recording devices.

Labeling and splits. The diagnostic labels are harmonized to support downstream multi-class and multi-label tasks. We adopt a consistent 80\%, 10\%, and 10\% split for training, validation, and testing sets, respectively. To avoid information leakage, all ECG recordings from the same patient appear in only one of these sets.

\subsection{Implementation Details and Evaluation Metrics}

In our method, the ECG patch size is set to 75, resulting in 30 patches for each input ECG segment with 2,250 timesteps. 
The Encoder employs 10 transformer layers with a latent dimension of 256 and 4 attention heads. 
The Latent Decoder employs 8 transformer layers with a latent dimension of 256 and 4 attention heads. 
The Time Decoder employs 10 transformer layers with a latent dimension of 256 and 4 attention heads. In STDM, masking probabilities are set as \(P_{\text{time}} = 0.5\) and \(P_{\text{lead}} = 0.2\). In the Lead Drop setting, the number of retained leads \(k\) is set to 4. In contrastive loss, the temperature coefficient $\tau$ in Eq.~(\ref{eq:contrastive_loss}) is set to 0.2. The loss weights for contrastive learning and reconstructive learning are set to 1 and 1, respectively.

During pre-training stage, we train for 80 epochs with a batch size of 256. 
The AdamW optimizer is used with the weight decay of 0.01 and the learning rate of \(1.5\times10^{-4}\). 
A cosine annealing scheduler with 5 warm-up epochs is adopted. 
During fine-tuning, we train for 80 epochs with a learning rate of \(8\times10^{-5}\) and no warmup; other hyperparameters remain the same as in pre-training. We follow the implementation of most baseline methods in ST-MEM. For the experimental metrics of other baseline methods already reported in ST-MEM, we directly adopt the recorded results. For ST-MEM, BMIRC, and a few methods whose metrics have not been reported, we perform pretraining on the same datasets and conduct careful hyperparameter tuning and evaluation on the same downstream datasets. All experiments are conducted on four NVIDIA RTX 3090 GPUs.

To evaluate model performance, we report Accuracy (ACC), F1-score (F1), and the Area Under the Receiver Operating Characteristic curve (AUROC) across different ECG datasets.
For single-label datasets such as PTB-XL and ICBEB2018, each sample has only one label, and Accuracy is computed at the sample level, measuring the proportion of samples whose predicted class matches the ground truth. F1-score and AUROC are computed per class and then averaged (macro) to equally weight each class, accounting for class imbalance. For multi-label datasets such as Ningbo, each sample may have multiple labels, and Accuracy is computed at the label level (micro accuracy), treating each label independently. F1-score and AUROC are similarly computed per class and macro-averaged to ensure that each label contributes equally, accommodating both the multi-label nature and class imbalance of the dataset. For metrics requiring binary decisions (i.e., Accuracy and F1-score), predicted probabilities are converted into binary labels using a fixed threshold of 0.5.

\subsection{Comparison with Other Methods}

We compare our CoRe-ECG with a comprehensive range of baselines, including purely supervised learning (Random Init), which directly trains our backbone on downstream datasets, and various SSL paradigms based on contrastive or reconstructive objectives. For contrastive learning, we evaluate MoCo v3 \cite{moco}, which focuses on instance-level consistency after data augmentation, and CMSC \cite{cmsc}, which maximizes the similarity between different temporal segments of the same ECG recording to capture temporal invariance. In the realm of reconstructive learning, we consider MTAE and MLAE \cite{maefe}, which reconstruct masked temporal and spatial patches respectively. Furthermore, we include ST-MEM \cite{st-mem}, a spatio-temporal masked modeling framework that explicitly captures multi-dimensional dependencies via lead-specific embeddings, and the recent BMIRC \cite{bmirc} method, which introduces a frequency-spectrum modality and internal representation connections to alleviate the decoder's reconstruction burden, thereby encouraging the encoder to prioritize the acquisition of high-level discriminative features.

\begin{table*}[t]
    \renewcommand{\arraystretch}{1.2}
    \centering
    \caption{Performance comparison among our model and baseline methods on three ECG datasets. 
    The best results are highlighted in bold and the second-best results are underlined.}
    \label{tab:comparison}
    \vspace{4pt}
    \small

    \resizebox{\textwidth}{!}{
    \begin{tabular}{l ccc ccc ccc}
        \toprule
        \multirow{2}{*}{\textbf{Methods}} &
        \multicolumn{3}{c}{\textbf{PTB-XL}} &
        \multicolumn{3}{c}{\textbf{ICBEB2018}} &
        \multicolumn{3}{c}{\textbf{Ningbo}} \\
        \cmidrule(lr){2-4} \cmidrule(lr){5-7} \cmidrule(lr){8-10}
         & ACC$\uparrow$ & F1$\uparrow$ & AUROC$\uparrow$
         & ACC$\uparrow$ & F1$\uparrow$ & AUROC$\uparrow$
         & ACC$\uparrow$ & F1$\uparrow$ & AUROC$\uparrow$ \\
        \midrule

        Supervised   & 71.4 & 60.2 & 89.6 & 75.3 & 73.7 & 93.7 & 83.2 & 37.6 & 83.9 \\
        \rowcolor{gray!15}
        MoCo V3 \cite{moco}     & 79.9 & \underline{64.4} & 91.3 & 85.2 & \underline{83.8} & 96.7 & 95.1 & 41.7 & 89.1 \\
        CMSC \cite{cmsc}       & 72.4 & 51.0 & 87.7 & 73.6 & 71.7 & 93.8 & 90.3 & 35.6 & 83.2 \\
        \rowcolor{gray!15}
        MTAE \cite{maefe}      & 78.9 & 61.3 & 91.0 & 79.3 & 76.9 & 96.1 & 94.4 & 41.1 & 87.7 \\
        MLAE \cite{maefe}        & 80.2 & 62.5 & 91.5 & 83.4 & 81.6 & 97.3 & 95.7 & 41.3 & 89.6 \\
        \rowcolor{gray!15}
        ST-MEM \cite{st-mem}     & \underline{80.9} & 63.4 & \underline{92.9} & \textbf{87.2} & 82.0 & \underline{97.7} & \underline{97.2} & \underline{42.5} & \underline{90.8} \\
        BMIRC \cite{bmirc}  & 75.2 & 62.1 & 88.3 & 79.6 & 66.2 & 92.4 & 70.4 & 41.7 & 90.2 \\

        \midrule
        \rowcolor{gray!15}
        \textbf{CoRe-ECG} 
        & \textbf{81.4} & \textbf{69.5} & \textbf{94.2} 
        & \underline{86.9} & \textbf{84.0} & \textbf{97.9} 
        & \textbf{97.7} & \textbf{43.4} & \textbf{92.2} \\

        \bottomrule
    \end{tabular}
    }
\end{table*}

In this section, we evaluate the proposed CoRe-ECG against the above baselines. All reported results (with the exception of the purely supervised Random Init baseline) are obtained following the representative pretrain and finetune paradigm. Specifically, the models are first pre-trained on the large-scale MIMIC-IV-ECG dataset \cite{gow2023mimic} to learn general electrocardiogram representations. Subsequently, the pre-trained weights are transferred to three diverse downstream datasets: PTB-XL, ICBEB2018, and Ningbo. Rigorous data partitioning was executed to ensure that no subjects overlap between the pre-training and downstream phases, thereby eliminating any risk of data leakage.

Table \ref{tab:comparison} summarizes the quantitative performance. It is evident that our proposed CoRe-ECG consistently outperforms the competitive baselines across the majority of metrics. Notably, on the PTB-XL dataset, our model achieves a 6.1\% improvement in F1-score compared to the reconstructive ST-MEM and a 5.1\% improvement compared to the contrastive MoCo V3. This performance gain can be attributed to our hybrid strategy, which synergistically combines reconstructive and contrastive objectives to capture both local morphological details and global semantic invariance.

While reconstructive methods like ST-MEM achieve high accuracy on ICBEB2018 (87.2\%), our model demonstrates superior discriminatory power, as reflected by a higher AUROC (97.9\%) and F1-score (84.0\%). The substantial performance leap over the Supervised baseline (e.g., +4.6\% AUROC on PTB-XL) underscores the necessity of self-supervised pre-training for mitigating the labeling scarcity in specialized ECG diagnostic tasks.

\subsection{Effect of Lead Reduction}

Table \ref{tab:lead-ablation} summarizes the model performance under various lead configurations, including standard 12-lead, 6-lead (limb leads), and single-lead (Lead I) settings. Our proposed model consistently maintains robust performance across all scenarios, outperforming the ST-MEM baseline \cite{st-mem} in nearly every metric. 

\begin{table*}[t]
    \renewcommand{\arraystretch}{1.2}
    \centering
    \caption{Performance comparison under different numbers of leads across three ECG datasets. 
    The best results for each lead configuration are highlighted in bold.}
    \label{tab:lead-ablation}
    \vspace{4pt}
    \small

    \resizebox{\textwidth}{!}{
    \begin{tabular}{
        l 
        >{\raggedright\arraybackslash}p{2.2cm} 
        ccc 
        ccc 
        ccc
    }
        \toprule
        \multirow{2}{*}{\textbf{Leads}} & 
        \multirow{2}{*}{\textbf{Models}} & 
        \multicolumn{3}{c}{\textbf{PTB-XL}} & 
        \multicolumn{3}{c}{\textbf{ICBEB2018}} & 
        \multicolumn{3}{c}{\textbf{Ningbo}} \\
        \cmidrule(lr){3-5} \cmidrule(lr){6-8} \cmidrule(lr){9-11}
         & & ACC & F1 & AUROC 
         & ACC & F1 & AUROC 
         & ACC & F1 & AUROC \\
        \midrule

        \multirow{2}{*}{12} 
            & ST-MEM     & 80.9 & 63.4 & 92.9 & \textbf{87.2} & 82.0 & 97.7 & 97.2 & 42.5 & 90.8 \\
            & \cellcolor{gray!15} CoRe-ECG   
              & \cellcolor{gray!15} \textbf{81.4} 
              & \cellcolor{gray!15} \textbf{69.5} 
              & \cellcolor{gray!15} \textbf{94.2} 
              & \cellcolor{gray!15} 86.9 
              & \cellcolor{gray!15} \textbf{84.0} 
              & \cellcolor{gray!15} \textbf{97.9} 
              & \cellcolor{gray!15} \textbf{97.7} 
              & \cellcolor{gray!15} \textbf{43.4} 
              & \cellcolor{gray!15} \textbf{92.2} \\

        \midrule

        \multirow{2}{*}{6}  
            & ST-MEM     & 76.8 & 57.7 & 90.2 & 80.3 & 77.7 & 95.2 & 95.1 & 36.7 & 89.4 \\
            & \cellcolor{gray!15} CoRe-ECG   
              & \cellcolor{gray!15} \textbf{77.1} 
              & \cellcolor{gray!15} \textbf{60.7} 
              & \cellcolor{gray!15} \textbf{90.3} 
              & \cellcolor{gray!15} \textbf{81.0} 
              & \cellcolor{gray!15} \textbf{78.8} 
              & \cellcolor{gray!15} \textbf{96.3} 
              & \cellcolor{gray!15} \textbf{96.7} 
              & \cellcolor{gray!15} \textbf{38.1} 
              & \cellcolor{gray!15} \textbf{90.9} \\

        \midrule

        \multirow{2}{*}{1}  
            & ST-MEM     & 67.4 & 42.1 & 81.5 & 71.1 & 68.8 & 93.7 & 93.3 & 23.1 & 86.5 \\
            & \cellcolor{gray!15} CoRe-ECG   
              & \cellcolor{gray!15} \textbf{69.5} 
              & \cellcolor{gray!15} \textbf{47.9} 
              & \cellcolor{gray!15} \textbf{82.4} 
              & \cellcolor{gray!15} \textbf{72.0} 
              & \cellcolor{gray!15} \textbf{69.3} 
              & \cellcolor{gray!15} \textbf{93.9} 
              & \cellcolor{gray!15} \textbf{94.6} 
              & \cellcolor{gray!15} \textbf{25.8} 
              & \cellcolor{gray!15} \textbf{88.7} \\

        \bottomrule
    \end{tabular}
    }
\end{table*}

Specifically, while reducing the leads from 12 to 6 typically results in performance degradation for most models, our method limits the drop in AUROC to only 3.9\% on PTB-XL and 1.6\% on ICBEB2018, significantly smaller than the declines observed in the baselines. Even in the extreme single-lead setting, our model retains substantial discriminative capability, achieving AUROCs of 82.4\% and 88.7\% on the PTB-XL and Ningbo datasets, respectively. These results demonstrate that the integration of lead-indicator embeddings combined with our sophisticated masking strategy enables the model to effectively capture inter-lead correlations. Consequently, the encoder can infer missing spatial information from visible leads, ensuring high diagnostic reliability even in lead-deficient clinical or wearable scenarios.

\subsection{Effect of Reduced Fine-Tuning Data}

To evaluate the generalization capability and data efficiency of our model, we simulate data-scarce scenarios by fine-tuning on reduced subsets of the downstream training data. Specifically, we randomly sample 50\% and 5\% of the original training sets to assess model robustness in low data regimes, while maintaining consistent validation and test splits as described in Section~\ref{sec:datasets}. 

The experimental results in Table \ref{tab:data-efficiency} demonstrate that our model consistently outperforms the state-of-the-art ST-MEM \cite{st-mem} across all data scales. Notably, in extreme low-data conditions (5\% subset), our model maintains a competitive AUROC of 88.1\% on PTB-XL and 97.1\% on ICBEB2018, significantly exceeding the performance of the baseline. Furthermore, we observe that using only 50\% of the training data with our method yields results comparable to, or even exceeding, those achieved by ST-MEM using the full 100\% dataset on PTB-XL (93.3\% vs. 92.9\% AUROC). This superior performance in data constrained settings suggests that our synergistic pre-training stage effectively extracts highly transferable and information-rich ECG representations, thereby reducing the dependency on large scale labeled datasets for downstream task adaptation.

\begin{table*}[t] % 跨双栏，置顶
    \renewcommand{\arraystretch}{1.3}
    \centering
    \caption{Performance comparison under different training data ratios (100\%, 50\%, and 5\%) on three ECG datasets. 
    The best results for each data ratio are highlighted in bold.}
    \label{tab:data-efficiency}
    \vspace{4pt}
    \small

    \resizebox{\textwidth}{!}{
    \begin{tabular}{l >{\raggedright\arraybackslash}p{2.2cm} ccc ccc ccc}
        \toprule
        \multirow{2}{*}{\textbf{Data Ratio}} & 
        \multirow{2}{*}{\textbf{Models}} & 
        \multicolumn{3}{c}{\textbf{PTB-XL}} & 
        \multicolumn{3}{c}{\textbf{ICBEB2018}} & 
        \multicolumn{3}{c}{\textbf{Ningbo}} \\
        \cmidrule(lr){3-5} \cmidrule(lr){6-8} \cmidrule(lr){9-11}
         & & ACC & F1 & AUROC & ACC & F1 & AUROC & ACC & F1 & AUROC \\
        \midrule

        \multirow{2}{*}{100\%} 
            & ST-MEM     & 80.9 & 63.4 & 92.9 & \textbf{87.2} & 82.0 & 97.7 & 97.2 & 42.5 & 90.8 \\
            & \cellcolor{gray!15} CoRe-ECG   
              & \cellcolor{gray!15} \textbf{81.4} 
              & \cellcolor{gray!15} \textbf{69.5} 
              & \cellcolor{gray!15} \textbf{94.2} 
              & \cellcolor{gray!15} 86.9 
              & \cellcolor{gray!15} \textbf{84.0} 
              & \cellcolor{gray!15} \textbf{97.9} 
              & \cellcolor{gray!15} \textbf{97.7} 
              & \cellcolor{gray!15} \textbf{43.4} 
              & \cellcolor{gray!15} \textbf{92.2} \\

        \midrule
        \multirow{2}{*}{50\%} 
            & ST-MEM     & 79.4 & 61.2 & 92.7 & 84.7 & 81.6 & 97.3 & 95.3 & 36.9 & 86.7 \\
            & \cellcolor{gray!15} CoRe-ECG   
              & \cellcolor{gray!15} \textbf{80.5} 
              & \cellcolor{gray!15} \textbf{64.7} 
              & \cellcolor{gray!15} \textbf{93.3} 
              & \cellcolor{gray!15} \textbf{86.2} 
              & \cellcolor{gray!15} \textbf{83.3} 
              & \cellcolor{gray!15} \textbf{97.6} 
              & \cellcolor{gray!15} \textbf{96.5} 
              & \cellcolor{gray!15} \textbf{37.1} 
              & \cellcolor{gray!15} \textbf{88.3} \\

        \midrule
        \multirow{2}{*}{5\%}  
            & ST-MEM     & 74.5 & 53.4 & 86.4 & 81.7 & 80.9 & 95.4 & 93.0 & 27.1 & 82.0 \\
            & \cellcolor{gray!15} CoRe-ECG   
              & \cellcolor{gray!15} \textbf{76.0} 
              & \cellcolor{gray!15} \textbf{57.1} 
              & \cellcolor{gray!15} \textbf{88.1} 
              & \cellcolor{gray!15} \textbf{83.9} 
              & \cellcolor{gray!15} \textbf{82.3} 
              & \cellcolor{gray!15} \textbf{97.1} 
              & \cellcolor{gray!15} \textbf{93.7} 
              & \cellcolor{gray!15} \textbf{28.9} 
              & \cellcolor{gray!15} \textbf{82.7} \\

        \bottomrule
    \end{tabular}
    }
\end{table*}

\subsection{Ablation Studies}

\subsubsection{Contrastive vs. Reconstructive vs. CoRe Learning}

To validate the advantage of joint pre-training compared to utilizing either contrastive learning or reconstructive learning in isolation, we conduct an ablation study by training models with only a single objective enabled. Our findings, summarized in Table \ref{tab:ab-pretrain}, demonstrate that contrastive and reconstructive learning are mutually beneficial processes. 

Specifically, while the Contrastive Learning variant achieves competitive AUROC scores (e.g., 93.6\% on PTB-XL), the Reconstructive Learning approach generally exhibits superior performance in capturing local morphological details, resulting in higher F1-scores across all three datasets. However, our proposed CoRe Learning framework achieves the highest performance across all metrics, reaching an AUROC of 94.2\% on PTB-XL, 97.9\% on ICBEB2018, and 92.2\% on Ningbo. This indicates that the synergy between the two objectives allows the model to simultaneously learn global semantic invariance and precise spatio-temporal signal reconstruction. 

Furthermore, we observe that inappropriate data augmentation strategies can negatively impact the reconstruction task by introducing physiologically misleading artifacts during training. By balancing these objectives, our model alleviates the limitations of single-paradigm approaches, extracting more comprehensive and discriminative ECG representations.

\begin{table*}[t] % 跨双栏，置顶
    \renewcommand{\arraystretch}{1.3}
    \centering
    \caption{Ablation study of different pre-training paradigms across three datasets. CoRe Learning (Ours) combines contrastive and reconstructive objectives. The best results for each dataset are highlighted in bold.}
    \label{tab:ab-pretrain}
    \vspace{4pt}
    \small
    \setlength{\tabcolsep}{0pt}
    \begin{tabular*}{\textwidth}{@{\extracolsep{\fill}} l ccc ccc ccc @{}}
        \toprule
        \textbf{Method} &
        \multicolumn{3}{c}{\textbf{PTB-XL}} &
        \multicolumn{3}{c}{\textbf{ICBEB2018}} &
        \multicolumn{3}{c}{\textbf{Ningbo}} \\
        \cmidrule(lr){2-4} \cmidrule(lr){5-7} \cmidrule(lr){8-10}
         & ACC & F1 & AUROC & ACC & F1 & AUROC & ACC & F1 & AUROC \\
        \midrule
        Contrastive Learning      & 80.6 & 68.7 & 93.6 & 83.4 & 82.5 & 97.0 & 96.3 & 41.1 & 91.3 \\
        Reconstructive Learning   & 81.2 & 69.1 & 94.0 & 84.4 & 83.0 & 97.3 & 96.5 & 42.4 & 91.7 \\
        \textbf{CoRe Learning} & \textbf{81.4} & \textbf{69.5} & \textbf{94.2} & \textbf{86.9} & \textbf{84.0} & \textbf{97.9} & \textbf{97.7} & \textbf{43.4} & \textbf{92.2} \\
        \bottomrule
    \end{tabular*}
\end{table*}

\subsubsection{Effect of Data Augmentation and Masking Strategy}

To evaluate the individual contributions of our proposed Frequency Dynamic Augmentation (FDA) and Spatio-Temporal Dual Masking (STDM) strategy, we conducted an ablation study under four distinct conditions. The results, summarized in Table \ref{tab:ablation-components}, demonstrate that both components significantly enhance the model's capability to capture essential ECG features.

\begin{table*}[t] % 跨双栏，置顶
    \renewcommand{\arraystretch}{1.3}
    \centering
    \caption{Ablation study of the proposed STDM and FDA. The checkmark ($\checkmark$) indicates the component is included, and ($\times$) indicates it is not.}
    \label{tab:ablation-components}
    \vspace{4pt}
    \small
    \setlength{\tabcolsep}{0pt}
    \begin{tabular*}{\textwidth}{@{\extracolsep{\fill}} c c ccc ccc ccc @{}}
        \toprule
        \multirow{2}{*}{\textbf{STDM}} & \multirow{2}{*}{\textbf{FDA}} & 
        \multicolumn{3}{c}{\textbf{PTB-XL}} & 
        \multicolumn{3}{c}{\textbf{ICBEB2018}} & 
        \multicolumn{3}{c}{\textbf{Ningbo}} \\
        \cmidrule(lr){3-5} \cmidrule(lr){6-8} \cmidrule(lr){9-11}
         & & ACC & F1 & AUROC & ACC & F1 & AUROC & ACC & F1 & AUROC \\
        \midrule
        $\times$ & $\times$ & 80.0 & 66.1 & 92.3 & 83.7 & 81.4 & 96.4 & 94.1 & 41.3 & 89.7 \\ 
        $\times$ & $\checkmark$ & 80.1 & 66.4 & 92.6 & 84.1 & 82.3 & 96.7 & 94.8 & 41.7 & 90.0 \\ 
        $\checkmark$ & $\times$ & 80.6 & 67.8 & 93.7 & 84.9 & 82.8 & 97.1 & 96.1 & 42.3 & 90.7 \\ 
        $\checkmark$ & $\checkmark$ & \textbf{81.4} & \textbf{69.5} & \textbf{94.2} & \textbf{86.9} & \textbf{84.0} & \textbf{97.9} & \textbf{97.7} & \textbf{43.4} & \textbf{92.2} \\ 
        \bottomrule
    \end{tabular*}
\end{table*}

Specifically, the introduction of frequency domain augmentation FDA alone yields consistent performance gains across all datasets (e.g., AUROC on PTB-XL increases from 92.3\% to 92.6\%), highlighting its ability to generate physiologically meaningful variations without introducing misleading artifacts. The STDM strategy provides an even more substantial improvement, raising the AUROC on the Ningbo dataset from 89.7\% to 90.7\%. This suggests that increasing the complexity of the reconstruction task effectively compels the encoder to model intricate inter-lead spatial correlations. Ultimately, the synergy between both components achieves the optimal performance, reaching peak AUROC scores of 94.2\%, 97.9\%, and 92.2\% on PTB-XL, ICBEB2018, and Ningbo, respectively.

\subsubsection{Analysis of Mask Rates in STDM}

\begin{figure}[t]
  \centering
  {\includegraphics[width=1.0\columnwidth]{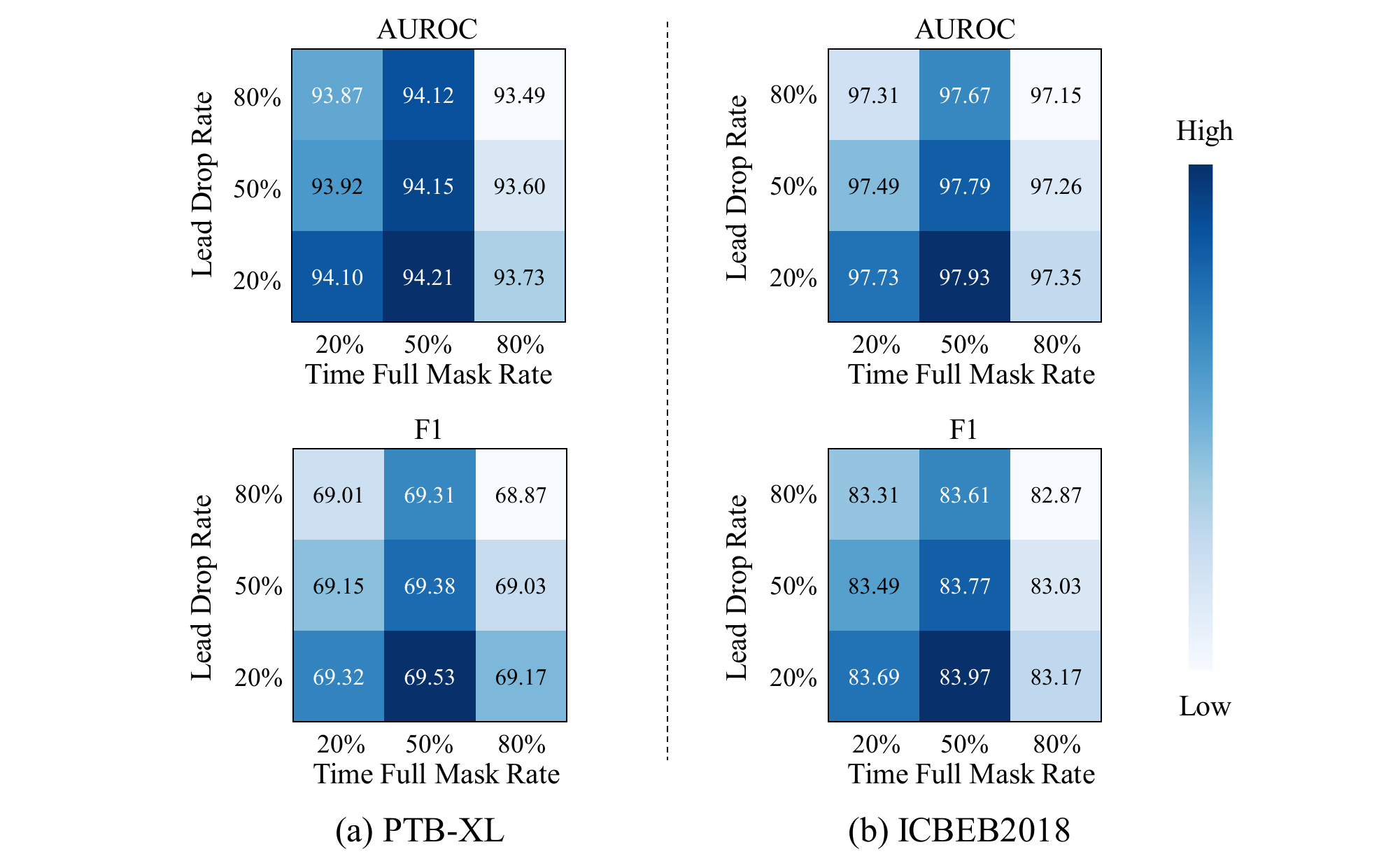}}
  \caption{Sensitivity analysis of the Time Full Mask Rate ($P_{\text{time}}$) and Lead Drop Rate ($P_{\text{lead}}$) in STDM strategy.}
  \label{fig:ab mask}
\end{figure}

The synergy between temporal and spatial masking in our STDM strategy is governed by the Time Full Mask Rate ($P_{\text{time}}$) and the Lead Drop Rate ($P_{\text{lead}}$). As illustrated in Fig. \ref{fig:ab mask}, the model performance across both PTB-XL and ICBEB2018 datasets exhibits a consistent parabolic trend relative to masking density. Specifically, CoRe-ECG achieves its peak diagnostic accuracy when $P_{\text{time}}$ is maintained at 0.5 and $P_{\text{lead}}$ at 0.2. We observe that an aggressive increase in $P_{\text{time}}$ to 0.8 precipitates a substantial performance decline, which underscores that obliterating entire temporal segments across all 12 leads renders the reconstruction task prohibitively difficult, thereby depriving the encoder of the essential rhythmic continuity required to extract stable physiological features. 

Furthermore, under a constant $P_{\text{time}}$, a higher $P_{\text{lead}}$ consistently correlates with diminished F1-scores and AUROC values. This phenomenon suggests that while removing spatial redundancies is necessary to prevent trivial solutions, an excessive drop rate starves the model of the inter-lead correlations defined by cardiac electrophysiology, ultimately hindering the formation of a robust latent representation. These empirical findings confirm that the optimal masking configuration is not merely about increasing task difficulty, but rather finding a critical threshold that forces the model to capture deep spatio-temporal dependencies without destabilizing the reconstructive learning process.

\subsection{Visualization analysis}

\begin{figure}[t]
  \centering
  \includegraphics[width=1.0\columnwidth]{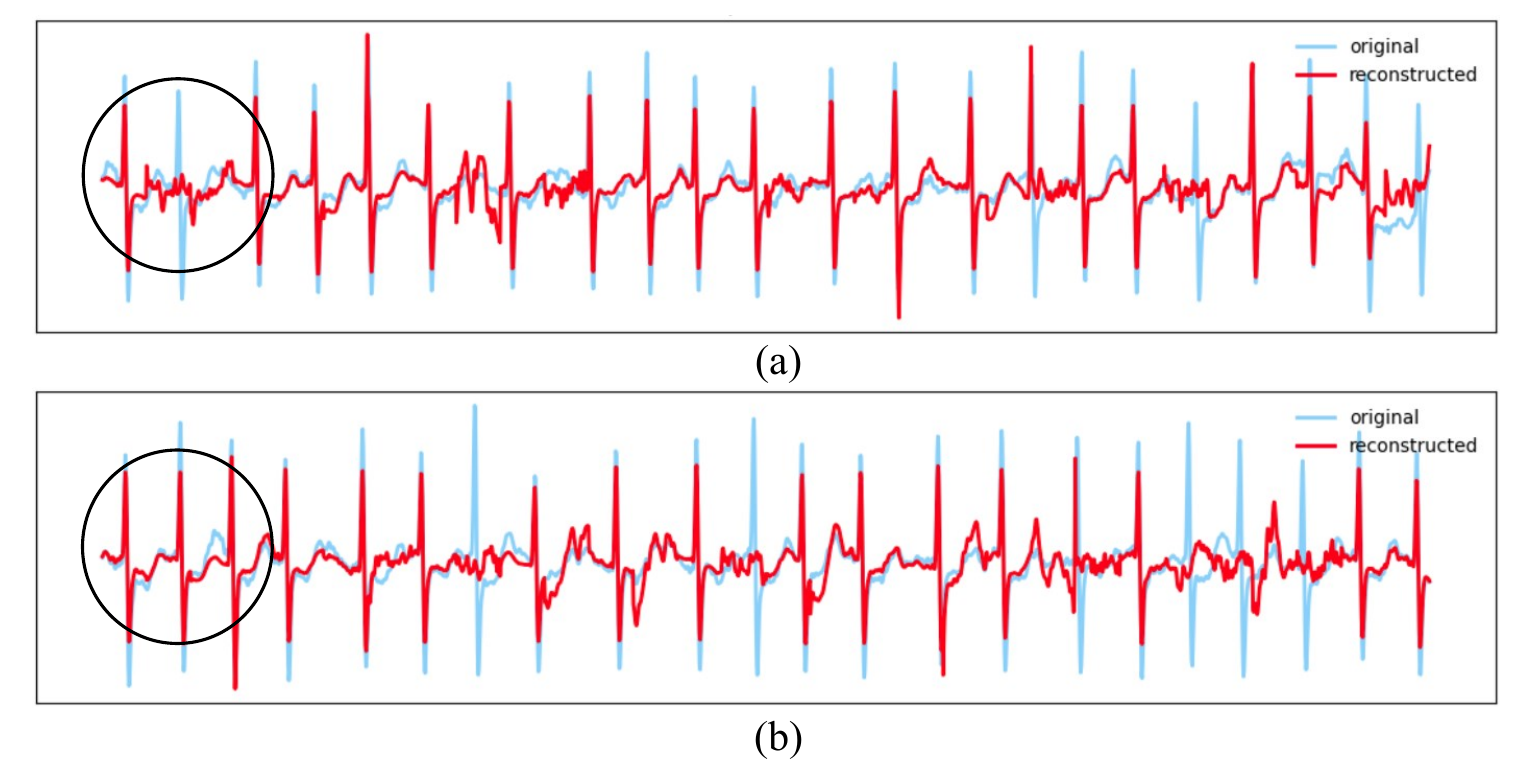}
  \caption{Visual comparison of reconstruction fidelity between ST-MEM (a) and our method (b). Red lines represent the reconstructed signals, while blue lines denote the original signals.}
  \label{fig:reconstruction}
\end{figure}

\begin{figure}[t]
  \centering
  \includegraphics[width=\columnwidth]{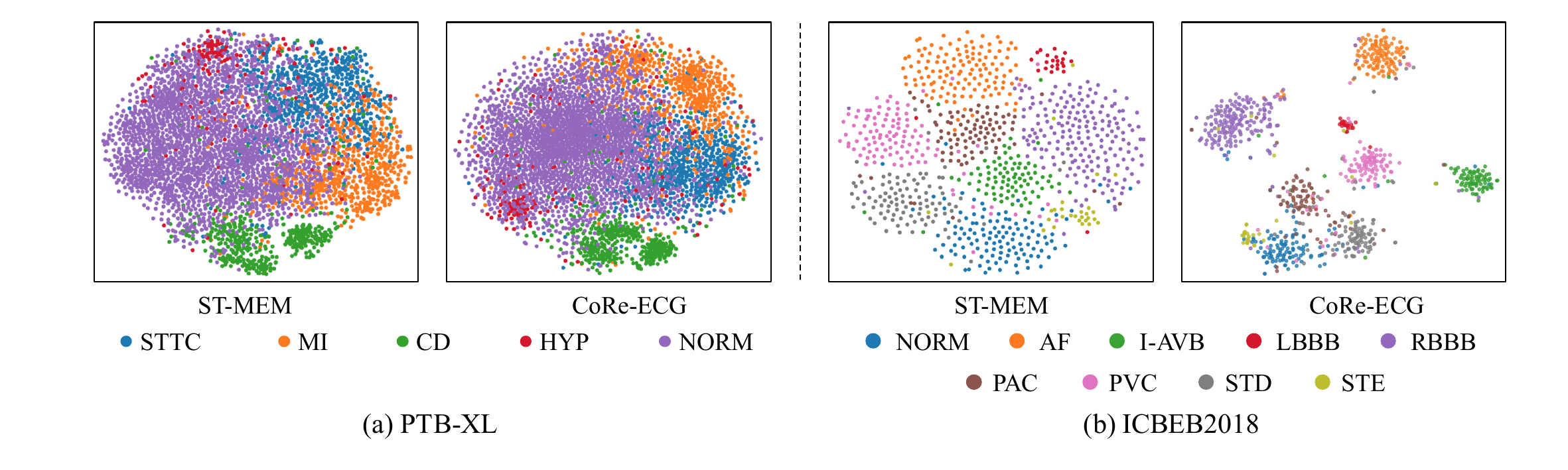}
  \caption{T-SNE visualization of the learned representations fine-tuned on two datasets. ICBEB2018 with 9 diagnostic categories. PTB-XL with 5 diagnostic categories.}
  \label{fig:tsne}
\end{figure}

To provide a more intuitive comparison between the baseline and the proposed method, we conduct a qualitative analysis focusing on signal reconstruction fidelity and feature space distribution.

% 重建图部分
As depicted in Fig.~\ref{fig:reconstruction}, which displays the reconstruction results for a representative ECG segment, our method significantly outperforms the baseline in recovering fine-grained morphological details. While the baseline tends to lose the precise shapes of the P-wave and T-wave components, our approach achieves high fidelity, suggesting that the STDM strategy successfully forces the model to capture subtle spatio-temporal dynamics essential for cardiac diagnosis.

% T-SNE 部分
Furthermore, we utilize T-SNE to project the high dimensional features from the ICBEB2018 dataset into a two dimensional space. As shown in Figs.~\ref{fig:tsne}, our method generates a more structured and separable feature distribution compared to the baseline. The distinct boundaries between different categories and the increased intra class compactness indicate that our model learns more discriminative representations, which facilitates downstream classification tasks.

% \begin{figure}[t]
%   \centering
%  \includegraphics[width=1.0\columnwidth]{tsne-ptb.pdf}
%   \caption{T-SNE visualization of the learned representations fine-tuned on the PTB-XL dataset. (a) shows the results of ST-MEM, while the (b) displays the results of our CoRe-ECG. Different colors represent the 5 diagnostic categories.}
%   \label{fig:tsne-ptb}
% \end{figure}


\section{Conclusion}
In this paper, we propose CoRe-ECG, a unified self-supervised pretraining framework that synergizes contrastive and reconstructive learning paradigms. Unlike existing methods that rely on a single learning objective, CoRe-ECG enables instance-level discriminative signals from contrastive learning to guide local waveform reconstruction, fostering cooperative learning of global semantic invariance and fine-grained spatio-temporal structural features of ECG signals. To further enhance the physiological rationality, we design Frequency Dynamic Augmentation (FDA) to perform adaptive frequency domain perturbations that preserve diagnostically critical ECG characteristics, and Spatio-Temporal Dual Masking (STDM) to break trivial linear dependencies across leads, eliminating shortcut learning in multi-lead ECG reconstruction. Extensive experiments on three public downstream ECG datasets demonstrate that CoRe-ECG achieves state-of-the-art performance. Ablation studies further validate the complementarity of contrastive and reconstructive learning, as well as the indispensable role of FDA and STDM in boosting model performance. Additionally, CoRe-ECG exhibits strong robustness in lead reduction and low data fine-tuning scenarios, verifying the high transferability of the learned ECG representations and its potential for clinical applications with limited labeled data or incomplete lead collection.

% \appendix
% \section{Example Appendix Section}


\clearpage 

% \bibliographystyle{elsarticle-num}
% \bibliography{cas-refs.bib}

% \end{thebibliography}
\end{document}